\documentclass[letterpaper, 10 pt, conference]{Resources/IEEEconf}  

\IEEEoverridecommandlockouts                              

\usepackage{etoolbox}
\makeatletter
\patchcmd{\@makecaption}
  {\scshape}
  {}
  {}
  {}
\makeatletter
\patchcmd{\@makecaption}
  {\\}
  {.\ }
  {}
  {}
\makeatother

\usepackage{gensymb}
\usepackage{graphicx}
\let\labelindent\relax 
\usepackage{lipsum}  

\usepackage{hyperref}
\usepackage{enumitem}
\usepackage{xcolor}
\usepackage{array}
\newcolumntype{x}[1]{>{\centering\let\newline\\\arraybackslash\hspace{0pt}}p{#1}} 
\usepackage{svg}
\usepackage{amsmath}
\svgpath{{../Figures/}}
\usepackage{float} 
\usepackage{algorithm}
\usepackage{wrapfig}
\usepackage[noend]{algpseudocode}

\title{\LARGE \bf
The Grasp Reset Mechanism: An Automated Apparatus for Conducting Grasping Trials
}
\author{
Kyle DuFrene\textsuperscript{1}, Keegan Nave\textsuperscript{1}, Joshua Campbell\textsuperscript{2}, Ravi Balasubramanian\textsuperscript{1}, and Cindy Grimm\textsuperscript{1}
\thanks{\textsuperscript{1}Collaborative Robotics and Intelligent Systems (CoRIS) Institute, Oregon State University, Corvallis, OR 97331 
\newline\indent\textsuperscript{2}Southwest Research Institute (SwRI), San Antonio, TX 6220
\newline\indent Thank you to our partners at the UMass Lowell Nerve Center, Brian Flynn, Adam Norton, and Holly Yanco.
\newline\indent This work was supported in part by the NSF under Grants CCRI 1925715 and RI 1911050.}}

\begin{document}
\maketitle
\thispagestyle{empty}
\pagestyle{empty}

\begin{abstract}
Advancing robotic grasping and manipulation requires the ability to test algorithms and/or train learning models on large numbers of grasps. Towards the goal of more advanced grasping, we present the Grasp Reset Mechanism (GRM), a fully automated apparatus for conducting large-scale grasping trials. The GRM automates the process of resetting a grasping environment, repeatably placing an object in a fixed location and controllable 1-D orientation. It also collects data and swaps between multiple objects enabling robust dataset collection with no human intervention. We also present a standardized state machine interface for control, which allows for integration of most manipulators with minimal effort. In addition to the physical design and corresponding software, we include a dataset of 1,020 grasps. The grasps were created with a Kinova Gen3 robot arm and Robotiq 2F-85 Adaptive Gripper to enable training of learning models and to demonstrate the capabilities of the GRM. The dataset includes ranges of grasps conducted across four objects and a variety of orientations. Manipulator states, object pose, video, and grasp success data are provided for every trial. 
\end{abstract}

\section{Introduction}

While humans have long mastered the art of grasping, autonomous robotic grasping and manipulation is an ongoing research field. Today, grasping in known, structured environments (such as loading CNC machines) is mostly a solved problem. However, grasping in less structured environments, with uncertainty about both the object type and its pose, still presents a difficult perception and decision-making problem. In response to these issues, advanced grasping algorithms and machine learning techniques have become ubiquitous in the field of robotic grasping. 

Machine learning models require large amounts of data to train. Successful machine learning models in the field of grasping frequently range from thousands to hundreds of thousands of grasps~\cite{li2019review}. Collecting this data in the real world requires costly and labor-intensive manual resetting of the setup. 

Alternatively, simulations are used to collect data and train machine learning models. Simulations provide the ability to run large amounts of grasping trials in a small period of time, and have near-infinite flexibility. However, simulations often fail to perfectly model the real world, leading to performance degradation when applied to real-world tasks (known as the Sim2Real gap~\cite{sim2real}). 

\begin{figure}
    \centering
    \includegraphics[width=.48\textwidth]{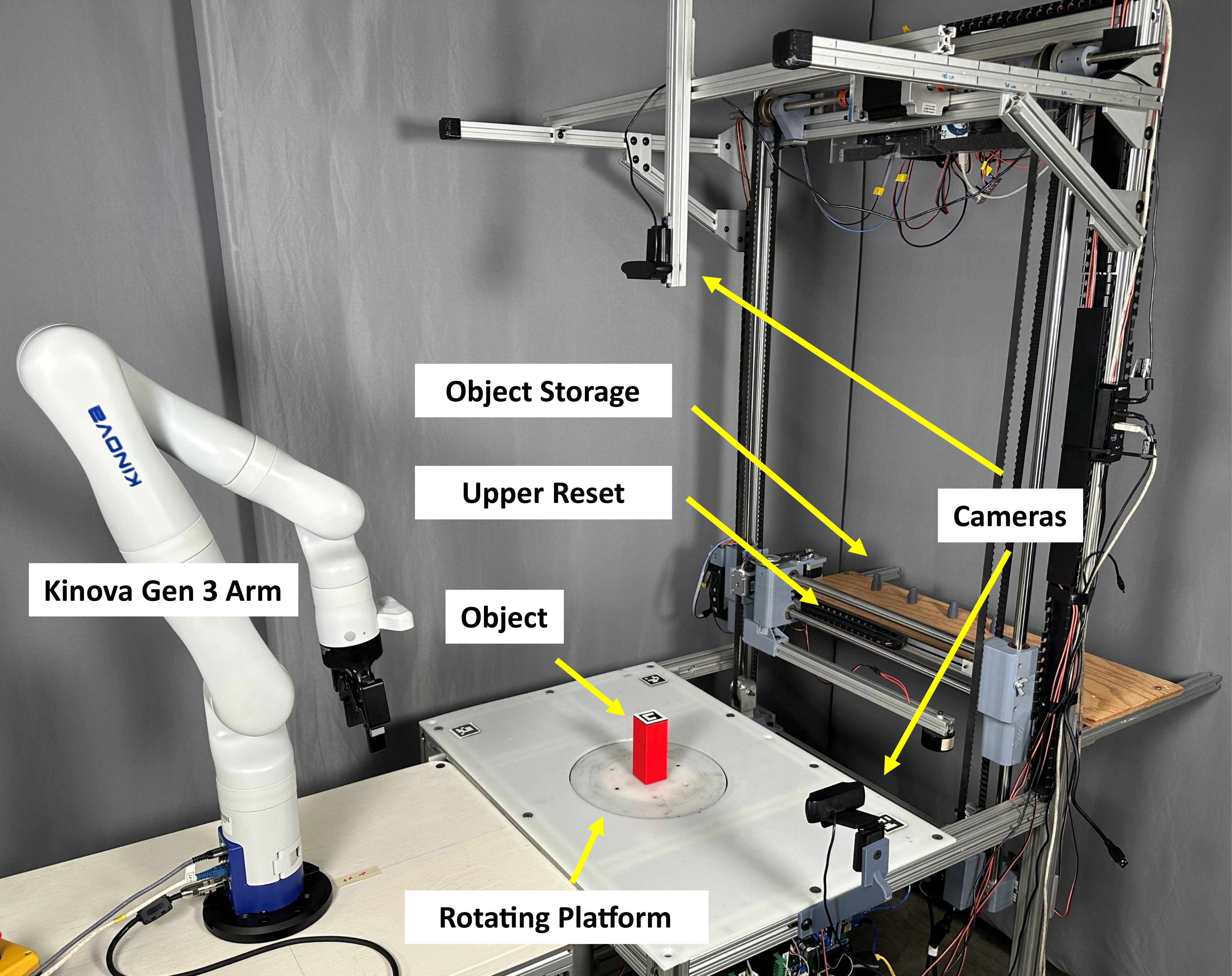}
    \caption{The Grasp Reset Mechanism with a Kinova Gen 3 robot attached. Visible components of interest are labeled.}
    \label{fig:grm}
\end{figure}

We have developed the Grasp Reset Mechanism~(GRM) in order to bridge the gap between inaccurate simulations and labor-intensive single-use test setups~(see Figure~\ref{fig:grm}). The GRM is a fully automated mechanism that precisely resets the environment for each grasping trial, integrates with external robots, and records data. It additionally is compatible with a wide variety of objects, and can even switch objects without human intervention. It is well suited to conduct large numbers of grasping or manipulation trials, validate grasping algorithms, benchmark manipulators, and create datasets.

This paper presents the mechanism design, the software, and various other components used to run the experiments. To demonstrate the capabilities of the GRM, we also present a dataset of 1,020 grasps conducted with a Robotiq 2F-85 Adaptive Gripper fixed to a Kinova Gen 3 robot. We conducted the grasps on four shapes representative of a wide variety of day-to-day objects --- a rectangular prism, triangular prism, cylinder, and cone. For each object, we captured grasps by varying the end-effector's pose relative to the object for several possible grasps. We focused on ``edge-regions'', poses where the grasp transitions from successful to failed.

\noindent {\bf Contributions:} Our Grasp Reset Mechanism and corresponding dataset provide valuable assets to the robotics community, including:
\begin{itemize}
    \item a hardware platform for automated grasp trials, the Grasp Reset Mechanism (GRM).
    \item software packages for conducing grasp trials, including the modular arm control implementation.
    \item a dataset of 1,020 grasps conducted on four different objects.
\end{itemize}

All aspects of the GRM and dataset discussed in this paper are open source and are available at \href{https://osurobotics.github.io/Physical-Robotic-Manipulation-Test-Facility/}{https://osurobotics.github.io/Physical-Robotic-Manipulation-Test-Facility/}.

\section{Related Work}
We broadly characterize robotic grasping strategies into two categories: analytical and empirical (also known as data-driven) approaches~\cite{overview}. In the past, the field of robotic grasping largely fell into the analytical category, with approaches like force-closure or form-closure. For example, Danielczuk et al. presented the Reach Model, which they validated with 2,625 real-world grasps in a human-reset and controlled environment~\cite{reach}. Because the tests require manual resets, this data set required constant human involvement. 

Recently, the focus has shifted to empirical, or data-driven, approaches. Due to the need for large numbers of grasps, many approaches using empirical techniques train networks in simulations for the speed and flexibility. The trained networks are then validated with a small number of real world grasps. An example is Iqbal et al.'s work presenting a deep reinforcement learning approach with a double deep Q-network (DDQN) to grasp various objects~\cite{baxter}. The author's approach of training in simulation and transferring to the real world showed reasonable overall performance, but there were still significant gaps in real world performance on certain tasks. Other works have presented similar approaches, but all suffer from the simulation to real (Sim2Real) gap~\cite{jacquard}. 

To circumvent the Sim2Real barrier entirely, some learning networks are trained with real world grasping attempts. A notable example is Levine et al.'s work presenting an approach to learn hand-eye coordination with 800,000 real world grasp attempts. These grasps were conducted over the course of two months with 14 manipulators~\cite{google}. Similarly, other works have trained models with 23,000, 25,000, 50,000 and 580,000 real world grasp attempts, respectively~\cite{self_supervised, self_supervised_2, supersizing, qt-opt}. However, as highlighted by Kleeberger et al. in their review, these approaches are not flexible to hardware changes (like a different gripper)~\cite{kleeberger_survey}. Notably, all of these are bin-picking tasks that use an image and a reduced pose set (2.5D) and a simple parallel-jaw style gripper.

For training learning networks, existing datasets of physical grasps are valuable. However, most datasets in the field of robotic grasping cover only part of the grasping problem. For example, some works provide samples of objects~\cite{ycb}, others localize objects~\cite{6d}, and others detect potential grasps from images~\cite{jacquard}. Only a few works have presented large-scale datasets of physical grasps, such as~\cite{google, supersizing}, but as mentioned previously they have limitations. 

\begin{figure}
    \centering
    \includegraphics[width=0.48\textwidth]{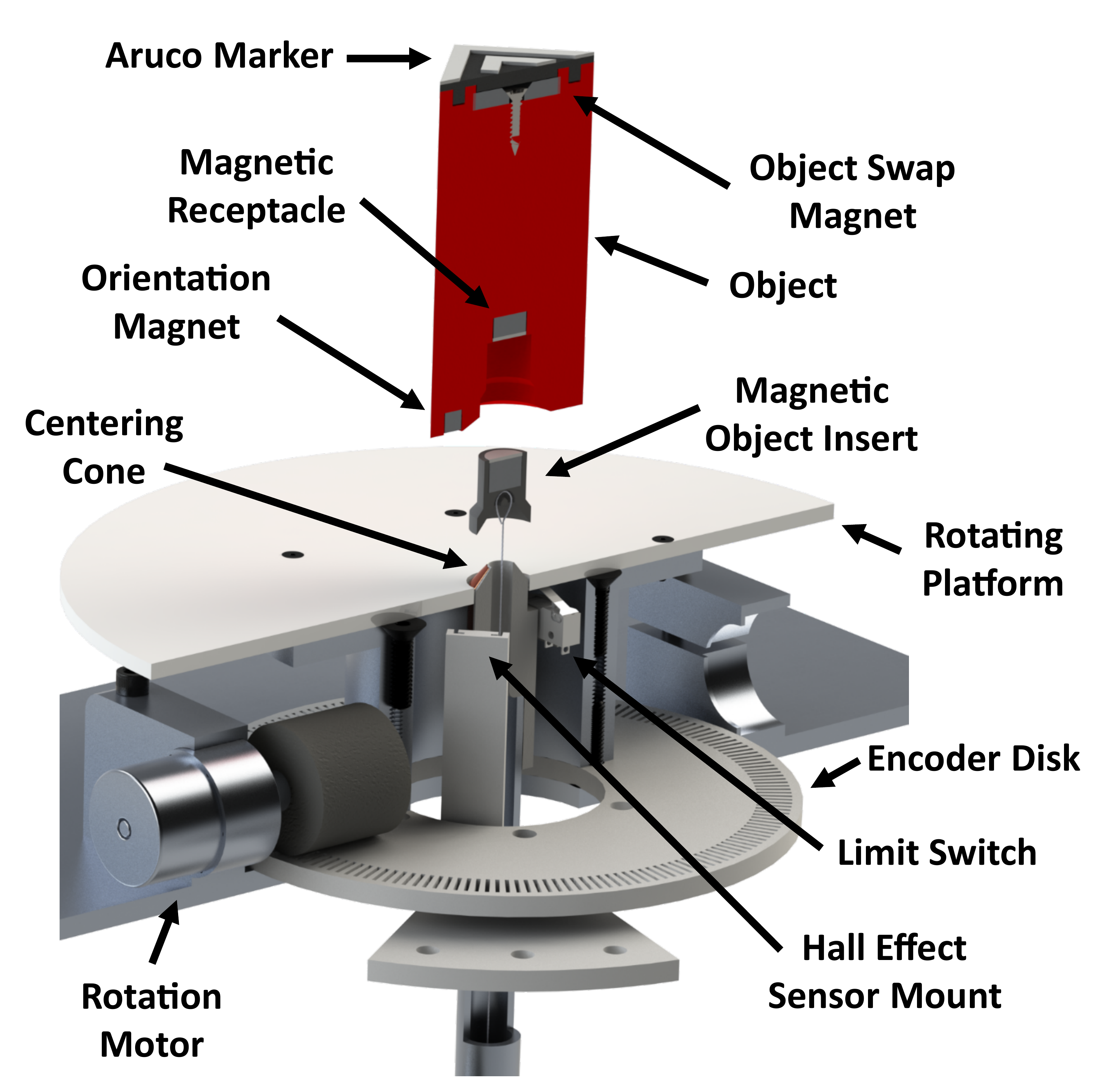}
    \caption{A cut-away view of the centering cone and rotation mechanism of the lower reset. Critical components are labeled. A rectangular prism is pictured as an example object.}
    \label{fig:cone}
\end{figure}
The GRM is unique in its ability to automatically create grasping datasets like the one presented in this paper. To date, we are not aware of any advanced, fully automated apparatuses like the Grasp Reset Mechanism used for grasping. We only know of limited examples that use similar mechanical principles. For example, Liarokapis et al. designed an object resetting mechanism with a top-mounted string in their paper creating manipulation primitives~\cite{primitives}. While the top down string allows for consistent position placement, it does not control rotation and only works with side grasps.

\section{Grasp Reset Mechanism}
The Grasp Reset Mechanism's (GRM) main goals are to: automate the resetting process of a grasping test environment, automate data collection, and interface with a variety of manipulators. We had a variety of design challenges, including:
\begin{itemize}
    \item How to \textit{reliably} manipulate an object without interfering with the GRM?
    \item How to control the object's orientation?
    \item How to switch between objects without operator intervention?
    \item How to seamlessly integrate the grasping environment with various manipulators?
\end{itemize}
\begin{figure*}
    \centering
    \includegraphics[width=.7\textwidth]{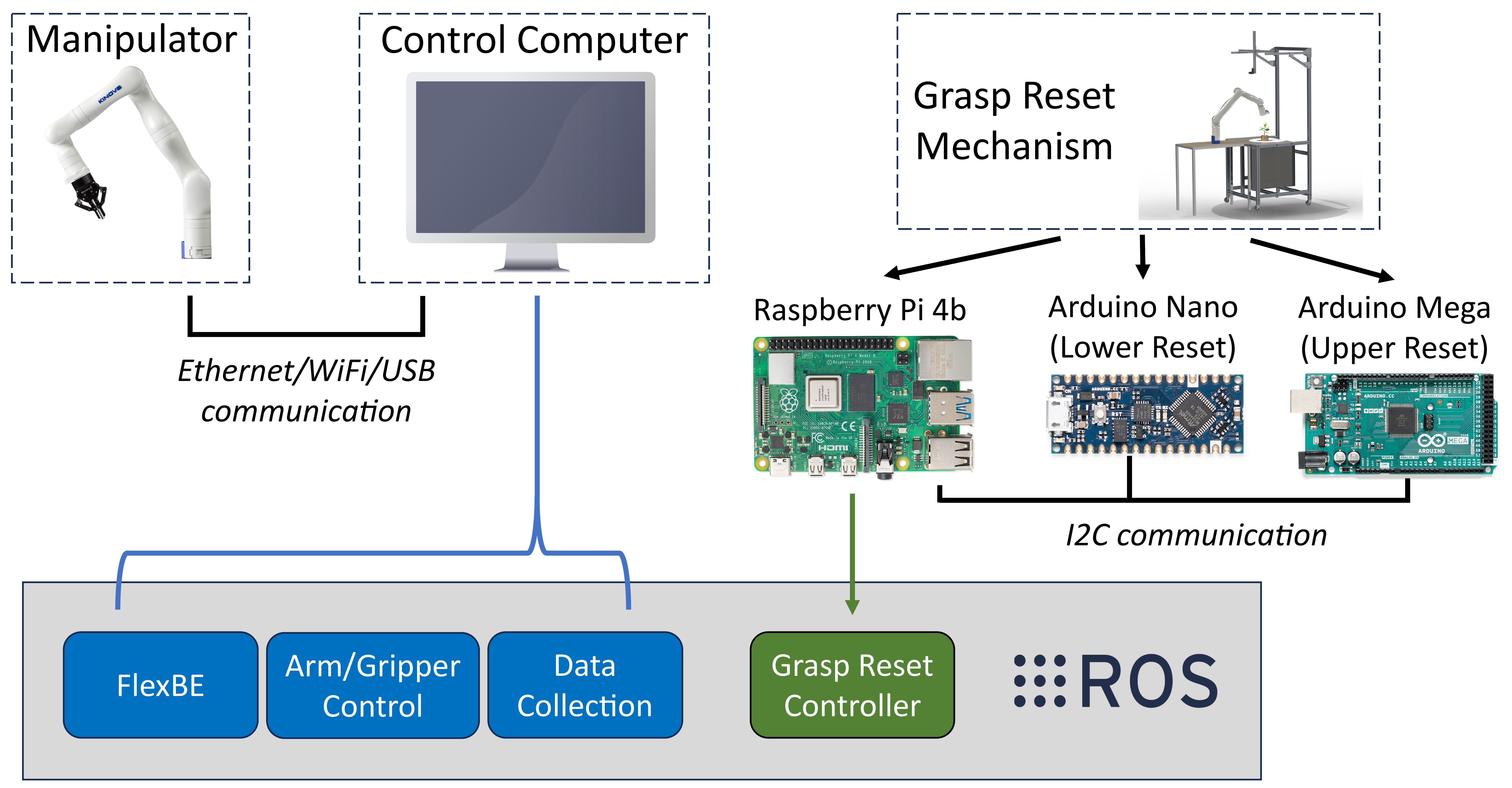}
    \caption{A high level view of control of the Grasp Reset Mechanism. The system includes three aspects, the control computer, arm, and GRM itself. The control computer interfaces with the arm via ethernet, WiFi, or USB and the Raspberry Pi on the GRM via ROS over WiFi/Ethernet.}
    \label{fig:flowchart}
\end{figure*}

These challenges led to our unique mechanical and software solutions. The GRM successfully resets the grasping environment, meaning the object returns to a fixed, home location {\em and} variable 1-D orientation. The GRM also supports automatically swapping between objects. Because we want to support grasp testing for a variety of arms/grippers and algorithms, we provide a standardized, state machine-based interface to schedule arm/gripper movements, collect data, and trigger the reset. 

The entire system design is broken into three components: mechanical, electrical, and software. We further discuss the compatibility of the GRM with other objects, as well as the reliability and accuracy.
\subsection{Mechanical Design}
The GRM is physically in two parts --- the lower and the upper reset. The lower reset provides the primary functionality (object resetting and orientation), whereas the upper reset adds the automated ability to switch objects. 

\subsubsection{Lower Reset}
The lower reset has three primary components: a retractable string attached to a magnetic object inserted in the base of the object, a centering cone, and a rotating platform. In a standard reset operation, the centering cone raises slightly above the level of the table. The string is then retracted, pulling the object on top of the centering cone. The string is then released and the centering cone lowered, resting the object on the rotating platform. The rotating platform then rotates to the desired target angle of the object. 

The centering cone is actuated by a NEMA stepper attached to a vertical ballscrew assembly. A limit switch attached to the lift mechanism identifies the lifted position. On each side of the cone are two copper plates that serve as a limit switch for string retraction. When the conductive magnetic object insert touches the two copper plates, an electrical short indicates the object is at its home position.

Fixed to one end of the string, the magnetic object insert, seen in Figure~\ref{fig:cone}, attaches to the magnetic receptacle of the object. This connection allows for simple object swapping (discussed in the next section). The other end of the string is a NEMA stepper, with a series of pulleys and a tensioner between. The tensioner is a simple gravity powered slotted tensioner, with a 70 gram weight. At the start of the trial, the stepper unwinds and the tensioner allows for the object to move up to 50 cm away.

The rotation mechanism rotates a 25 cm platform around the centering cone. A hall effect sensor fixed underneath the rotation mechanism detects a small magnet located off-center on the bottom of the object to determine the absolute starting orientation of the object. A 3D printed encoder disk and optical endstop function as an encoder for precise rotation angles. The centering cone and rotation mechanism are shown in Figure~\ref{fig:cone}.

As the string is the sole physical connection between the object and the GRM, the lower reset mechanism does not interfere with top or side grasps and most manipulation movements.

\subsubsection{Upper Reset}
The upper reset is a three degree of freedom arm mounted on the back of the GRM. The arm removes and replaces objects on the lower reset with a different, pre-loaded object on a shelf. This is known as an object swap. The arm can translate along the $x$-axis (left/right) and $z$-axis (up/down), and rotate about the $z$-axis of the GRM. 

In a normal swap operation, the lower reset first raises the centering cone and retracts the string. This fixes the magnetic object insert to the top of the centering cone. The upper reset arm then lifts up, rotates out 90 degrees, and moves to the middle directly above the object. The electromagnet at the end of the arm is powered, and the arm is raised up. The magnetic field generated by the electromagnet is stronger than that of the magnetic object insert, and the object is decoupled from the magnetic object insert in the base. The arm then places the object in the object storage area at the rear of the grasp reset mechanism, and selects and places a different object back on the magnetic object insert. 

All three axes are driven by NEMA steppers and have limit switches located at their home positions for determining absolute position upon startup.

\subsection{Electrical Design}
The electrical system is divided into two parts, largely matching the mechanical setup -- the upper and lower reset. A flowchart showing a high level overview of the software and electrical setup is shown in Figure~\ref{fig:flowchart}.

A Raspberry Pi serves as the main controller for the GRM, communicating via ROS with the control computer. The Raspberry Pi controls two microstepper drivers and an H-bridge on the lower reset, and communicates via I\textsuperscript{2}C with two Arduinos. The two Arduinos, an Arduino Nano on the lower reset and an Arduino Mega on the upper reset, monitor the limit switches with interrupts and control all other actuators on the GRM. Specifics about the lower and upper reset are presented below.

\subsubsection{Lower Reset}
The lower reset electrical panel includes the following:
\begin{itemize}
    \item 12 and 24 volt power supplies
    \item Microstepper drivers for the centering cone and string steppers
    \item Fuse box
    \item A custom PCB with:
    \begin{itemize}
        \item Raspberry Pi 4b
        \item Arduino Nano
        \item DC motor driver for the rotation mechanism
        \item 3.3 and 5 volt buck converters
        \end{itemize}
\end{itemize}

On the custom PCB, general-purpose input output (GPIO) pins on the Pi are connected directly to the microstepper drivers and DC motor driver for control of the motors. 
An additional breakout PCB adds an emergency stop switch. The emergency stop switches motor power off, but retains power to the Raspberry Pi and Arduino.

All motors, sensors, and external PCBs are connected via screw terminals or JST type connectors for quick, reliable connections and disconnections.

\subsubsection{Upper Reset}
The upper reset panel contains just five components, an Arduino Mega, an H-bridge, and three microstepper drivers. The Aduino Mega communicates with the Pi via I2C (also through a logic level shifter), and controls all aspects of the upper reset process. This includes the steppers, electromagnet, and monitoring limit switches. 12 and 24 volt power is provided to the upper reset by power supplies on the lower reset. 
\subsection{Software}
The overall framework of the software is the Robot Operating System (ROS)~\cite{ros}, with an external master control computer communicating with a peripheral Raspberry Pi located on the GRM. The master computer controls all aspects of trials, including the state machine, arm control, and data collection. The Raspberry Pi on the GRM only controls the physical reset operations based on control parameters passed from the control computer. ROS nodes on the control computer and the Raspberry Pi run within custom Docker containers for easy deployment to new devices.
\subsubsection{Control Computer}

\begin{figure}
    \centering
    \includegraphics[width=0.48\textwidth]{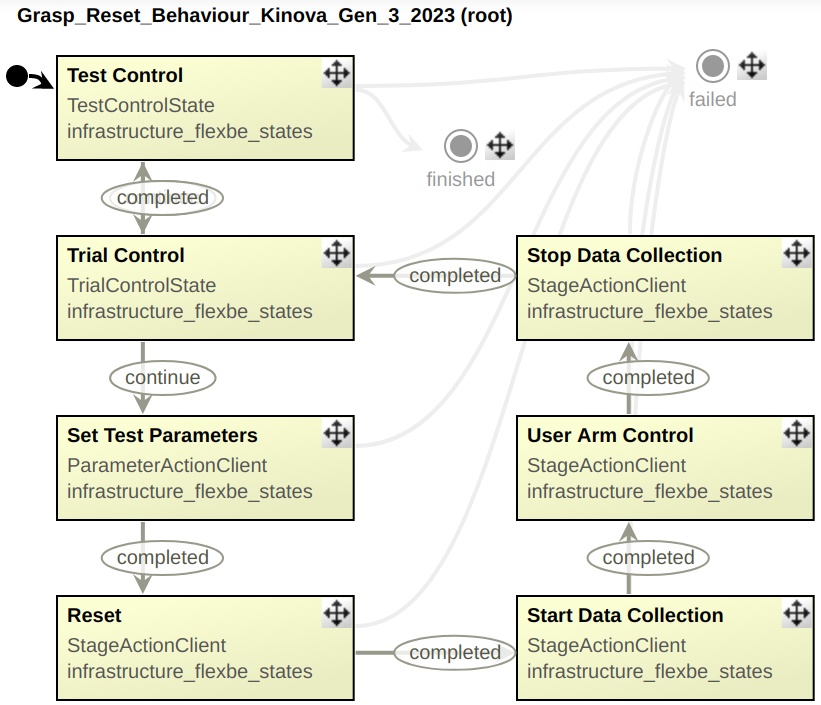}
    \caption{The FlexBE state machine showing trial progression. A series of trials starts with Test Control in the upper left, then moves down to Trial Control and repeats in a counter-clockwise pattern until all trials are complete (or a failure occurs).}
    \label{fig:flexbe}
\end{figure}

\noindent{\bf State Machine:} The hierarchical state machine controls the three high-level behaviors: resetting the environment, arm/gripper movements, and data collection. FlexBE was chosen as the state machine package due to its ROS integration and ease of use~\cite{flexbe}. At the beginning of a set of trials, the user loads a CSV containing information about the number of trials, requested objects, and what the object rotation angles are. All FlexBE actions are ROS action servers, which execute the corresponding action as demonstrated in Figure~\ref{fig:flexbe}. 

\noindent{\bf Arm Control:} Arm and gripper actions are prompted by the FlexBE state machine via a ROS action client. Because the action server topics remain constant, users may load custom manipulator movements matching any desired grasp tests. Upon completion, the action server returns a success or fail status and FlexBE responds accordingly. With this simple action client implementation, virtually any arm can interface with the GRM. This paper uses the Kinova Gen3 arm, although the Kinova Jaco 2 and Universal Robots arms have been validated with this system.

\noindent{\bf Data Collection:} The user can chose to turn data collection on (or not). When data collection is active, FlexBE will prompt a data collection action server which records various topics in a rosbag. For the trials within this paper, the arm and gripper states, the arm RGBD wrist camera, and a top and side RGB camera (mounted to the GRM) are recorded. Additional sensors, cameras, or states can easily be added by including their ROS topic in the rosbag.

\subsection{Compatibility}
\begin{figure}
    \centering
    \includegraphics[width=0.48\textwidth]{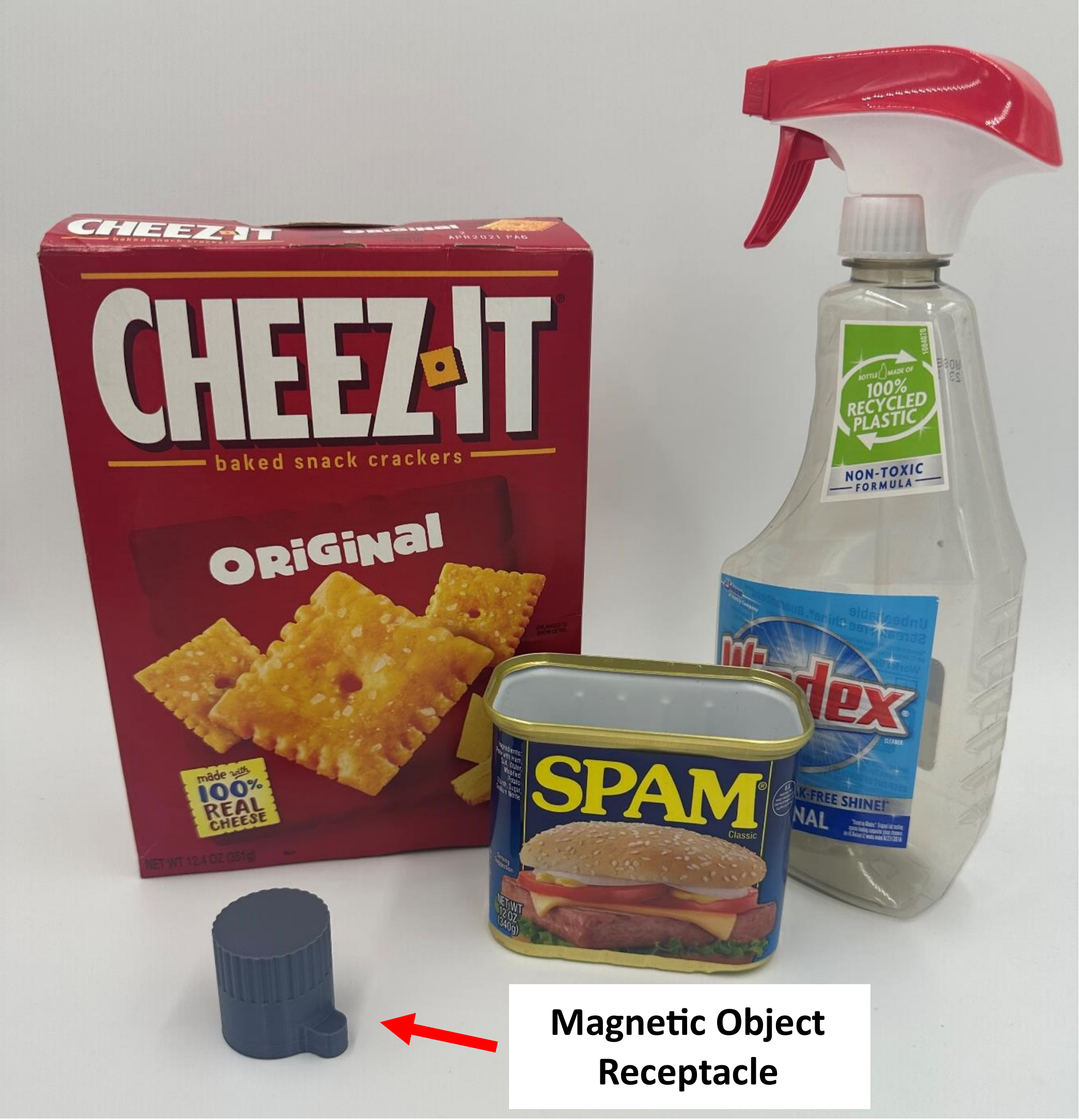}
    \caption{The magnetic object receptacle is pictured in the bottom left, with three objects from the YCB data set~\cite{ycb} --- a CheezIt box, SPAM can, and spray bottle. Using these objects with the GRM is straightforward, simply a hole must be cut in the bottom and the receptacle glued in.}
    \label{fig:cheezit}
\end{figure}

The GRM is compatible with a wide variety of objects. Any object that can have a receptacle for the bottom magnetic object insert and orientation magnet, and is under the size and weight limitations of the GRM may be used with the lower reset. Generally, any object less than 200 mm in all dimensions and 1 kilogram is compatible with the GRM. While only rigid object have been tested, semi-rigid/soft objects that hold their shape are also compatible. A variety of compatible objects from the YCB object set~\cite{ycb} and the magnetic receptacle are shown in Figure~\ref{fig:cheezit}. 

If automated object swapping by the upper reset is desired, there must be a larger magnet mounted at the top of the object. Object swapping does add additional limitations to maximum object size and weight --- 75 mm in width/depth and 500 grams.

Additionally, as discussed in the Software section, manipulators that can be controlled with a ROS action server may be used with the GRM. Manipulators that do not have a range of at least 50 cm may require special mounting/connection solutions (for example, a hand could be fixed directly in front of the object to test manipulation tasks).

\subsection{Repeatability and Accuracy}
As the GRM was designed to minimize human intervention and input, repeatability and accuracy are critical properties. We conducted two tests to quantify these. First, we conducted a series of 250 identical trials (moving the object and running the reset process). To determine repeatability, we monitored the mechanism for failures. Second, we performed 20 reset operations with the rectangular prism with an Aruco marker placed on top. Using images recorded from the top-mount camera, we tracked that Aruco marker's position and orientation at the end of each reset to determine variation in position and orientation.

For the 250 repeatability trials, no failures occurred --- no human intervention or supervision was required. 

In the 20 reset accuracy trials, the variation in starting position (planar, x-y) had a mean of $.05$ mm standard deviation of $.02$ mm. The variation in orientation had a mean of $2.0$ degrees standard deviation of $1.3$ degrees.
\begin{figure}
    \centering
    \includegraphics[width=0.47\textwidth]{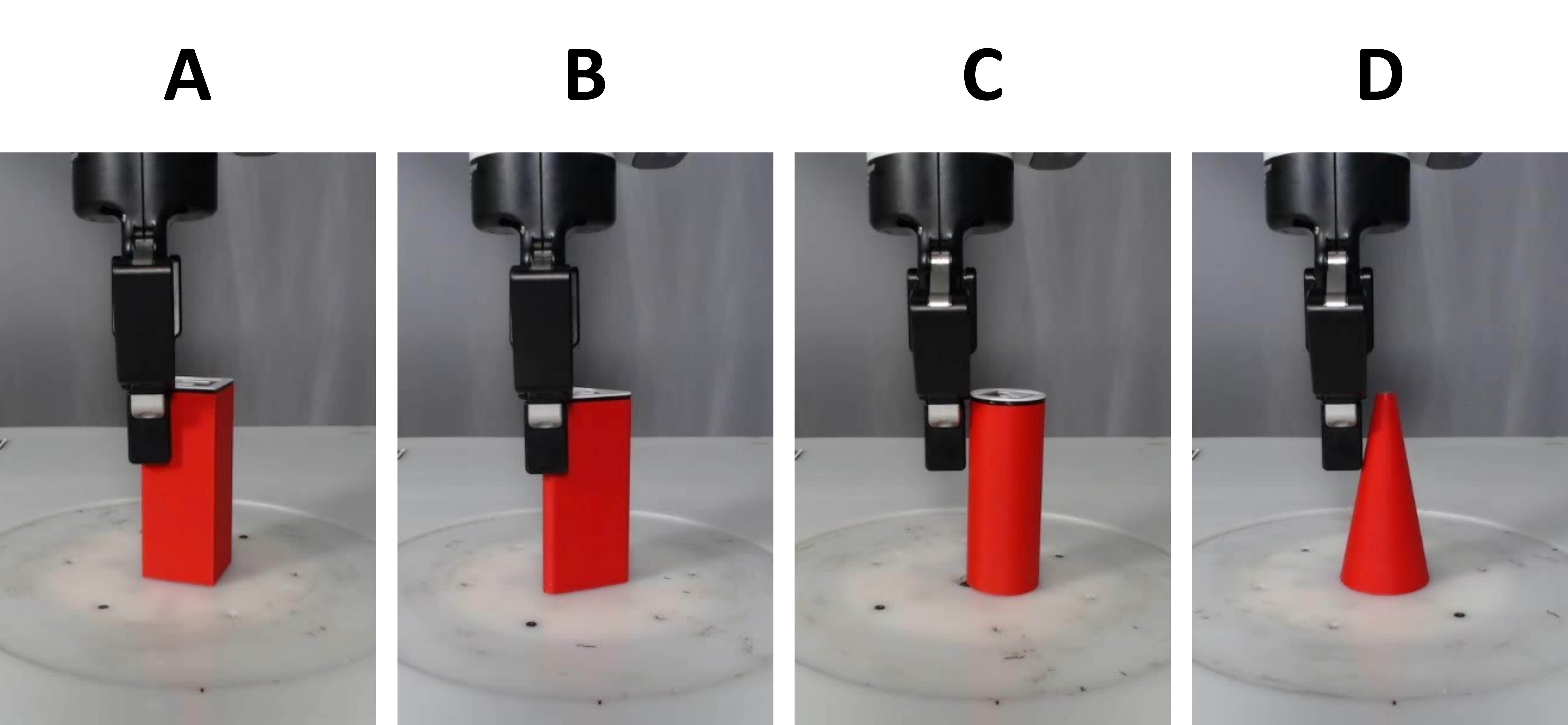} 
    \caption{Example grasp results on all four objects with the same manipulator pose. From left to right the objects are the rectangular prism (A, success), triangular prism (B, success), cylinder (C, failure), and cone (D, failure). The images are captured from the side mounted GRM camera.}
    \label{fig:success_fail}
\end{figure}

\section{Dataset}
We also provide a dataset with a wide variety of grasps across multiple objects. The following sections detail the specifics of the dataset and results of the grasp trials.

\subsection{Dataset Methods}
\begin{table*}
\caption{The table presents all combinations of grasps and their success rates in the dataset. The first two columns, ``End Effector Pose" and ``Range" are the type of deviation applied to the end effector and the range applied across the 15 trials. The ``Grasp Type" column notes whether the series of grasps were performed in a top down manner or from the side. In the remaining columns, the angles the GRM rotated the object to and corresponding overall grasp success rates are presented for each of the four objects.}
\begin{center}
\def\arraystretch{1.4}%
\begin{tabular}{ | m{5.5em} | m{4em} | m{2.5em} || m{5em} | m{3em} || m{5em} | m{3em} || m{5em} | m{3em} || m{5em} | m{3em} ||} 
  \hline
   \multicolumn{3}{|c||}{ } & \multicolumn{2}{c||}{\textbf{Rectangle}} & \multicolumn{2}{c||}{\textbf{Triangle}}& \multicolumn{2}{c||}{\textbf{Cylinder}} & \multicolumn{2}{c||}{\textbf{Cone}}\\
   \hline
  \textbf{End Effector Pose} & \textbf{Range} & \textbf{Grasp Type} & \textbf{Object \newline Angle (\degree)} & \textbf{Success Rate} & \textbf{Object \newline Angle (\degree)} & \textbf{Success Rate} & \textbf{Object \newline Angle (\degree)} & \textbf{Success Rate} & \textbf{Object \newline Angle (\degree)} & \textbf{Success Rate}\\ 
  \hline
  X translation & 0-3 cm & Top & 0, 15, 30, 45 & 68\% & 0, 20, 40, 60 & 35\% & 0 & 53\% & 0 & 40\%\\ 
  \hline
  Y translation & 1-5 cm & Top & 0, 15, 30, 45 & 73\% & 0, 20, 40, 60 & 60\% & 0 & 87\% & 0 & 80\%\\ 
  \hline
  X rotation & 0-45\degree & Top & 0, 15, 30, 45 & 65\% & 0, 20, 40, 60 & 45\% & 0 & 80\% & 0 & 47\%\\  
  \hline
  Y rotation & 0-90\degree & Top & 0, 15, 30, 45 & 95\% & 0, 20, 40, 60 & 90\% & 0 & 47\% & 0 & 33\%\\ 
  \hline
  Z rotation & 0-60\degree & Top & 0 & 93\% & 0 & 87\% & 0 & 100\% & 0 & 100\%\\ 
  \hline
  X translation & 0-5 cm & Side & 0, 15, 30, 45 & 55\% & 0, 20, 40, 60 & 37\% & 0 & 53\% & 0 & 47\%\\
  \hline
  X rotation & 0-45\degree & Side & 0, 15, 30, 45 & 100\% & 0, 20, 40, 60 & 97\% & 0 & 93\% & 0 & 100\%\\
  \hline
  Z rotation & 0-60\degree & Side & 0 & 100\% & 0 &  100\% & 0 & 100\% & 0 & 47\%\\ 
  \hline
\end{tabular}
\end{center}

\label{table:parameters}
\end{table*}

We collected a dataset of grasps using a Kinova Gen3 arm with a Robotiq 2F-85 gripper. Our goal was to find where grasps transition from successful (eg, grab the side of the cube with the cube centered in the gripper) to unsuccessful (the gripper is offset and knocks the cube over, or fails to get a proper grasp). This mimics the case where the object's pose relative to the gripper is noisy/incorrect. Four objects were used: a rectangular prism, a triangular prism, a cylinder, and a cone. Each of the objects has a top mounted Aruco marker~\cite{aruco}, except the cone with has a colored dot for object tracking due to the small diameter of its tip. Each object measured 40 mm (except the triangular prism, which measured 50 mm) in width/depth and was 105 mm tall. These basic shapes were chosen to represent a wide variety of objects in everyday life such as packaged food products and knobs.

The two prisms were tested at several orientations relative to the grasp direction (i.e., putting either the flat side or the angular side of the prism into the palm). The rectangular prism was tested at 0°, 15°, 30°, and 45°. The triangular prism was tested at 0°, 20°, 40°, and 60°. 

\begin{figure}
    \centering
    \includegraphics[width=0.48\textwidth]{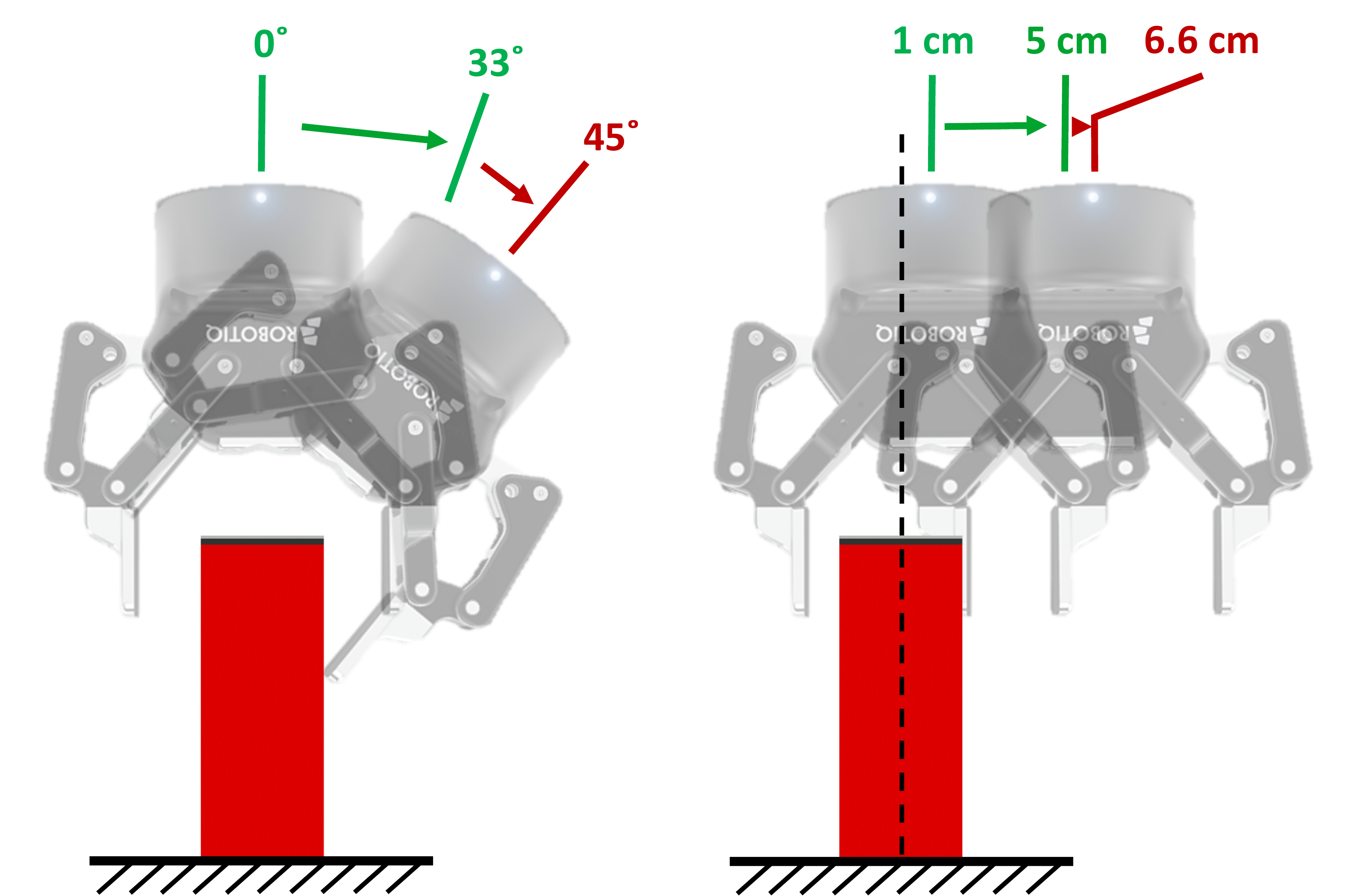}
    \caption{Two sets of end effector modifications are presented. In both, the grasps are from the top and the object is a rectangular prism rotated to 0\degree. The left image shows end effector $Y$-axis rotation, with success from 0 to 33\degree. All grasps failed between 33 to 45\degree. The right image shows end effector $Y$-axis translation, with success from 1 to 5 cm and failures between 5 cm to 6.6 cm.}
    \label{fig:dataset_example}
\end{figure}

For each combination above, 15 trials were conducted across a range of end effector rotations or translations. The 15 trials are equally spaced between the center of the grasp range for that axis to the edge. Examples of these end effector modifications are shown in~\ref{fig:dataset_example}. Generally, some grasps are expected to succeed and some fail for every object. All of the combinations of trials are presented in Table~\ref{table:parameters}.

Following the grasp, the object was lifted slightly and moved horizontally to a fixed target location 25 cm away from the center. Grasps are considered successful if the gripper position never reads fully closed.

\subsection{Dataset Results}
We captured 1,020 grasp trials in our dataset. Of those, 715 (70\%) were successful. More information about success rates for each object and end effector orientation are presented in Table~\ref{table:parameters}.  Each trial took approximately one minute, including manipulator planning, movement, and the resetting process, for a total of 17 hours.

The dataset includes the following for each trial:
\begin{itemize}
    \item Grasp pose
    \item Arm and gripper states (position and velocity)
    \item Object shape and rotation
    \item Grasp status (success/failure)
    \item Top and side view RGB camera feed (side view feed demonstrated in Figure~\ref{fig:success_fail})
    \item Wrist mounted RGBD camera feed
\end{itemize}

\addtolength{\textheight}{-13cm} 
\section{Discussion and Conclusions}
We presented the Grasp Reset Mechanism, an automated apparatus for conducting grasping trials and datasets. The GRM automatically resets an object at the end of a grasping trial, and moves it to a constant central position with a desired rotation (about its vertical axis). Our novel string and centering cone mechanism are robust and repeatable, and allow for a wide range of compatible objects. Additionally, the GRM can switch between multiple objects, allowing for hundreds of trials with varying objects and minimal human interaction. 

Our corresponding open source software package is also  designed for seamless integration with any manipulator. Automated data collection and post-processing, combined with the physical mechanism, further enhance the utility of the GRM. 

We also present a dataset that demonstrates the capabilities of the GRM. With 1,020 trials conducted over 17 hours and across four objects, the dataset may be used to train or validate machine learning models or other algorithms. The dataset also provides interesting insight into the robustness of certain grasping poses. For example, Table~\ref{table:parameters} demonstrates that side grasps with X or Z end effector modifications are significantly more robust than top grasps with the same modifications.

Beyond producing  large datasets for training, one potential future use is for benchmarking. Due to its repeatability and lack of required human involvement, manipulators or even grasping algorithms/models can be directly compared at scale.

We provide CAD files, electrical schematics, software, and the dataset on our website (provided at the end of the introduction).

\bibliographystyle{IEEEtran}
\bibliography{main}
\end{document}